# Exploiting Evidence in Probabilistic Inference

**Mark Chavira, David Allen, and Adnan Darwiche**
Computer Science Department
University of California
Los Angeles, CA 90095-1596
{chavira,dlallen,darwiche}@cs.ucla.edu

## Abstract

We define the notion of compiling a Bayesian network *with evidence* and provide a specific approach for evidence–based compilation, which makes use of logical processing. The approach is practical and advantageous in a number of application areas—including maximum likelihood estimation, sensitivity analysis, and MAP computations—and we provide specific empirical results in the domain of genetic linkage analysis. We also show that the approach is applicable for networks that do not contain determinism, and show that it empirically subsumes the performance of the quickscore algorithm when applied to noisy–or networks.

## 1 INTRODUCTION

It is well–known that exploiting evidence can make inference in a Bayesian network more tractable. Two of the most common techniques are removing leaf nodes that are not part of the evidence or query (Shachter, 1986) and removing edges outgoing from observed nodes (Shachter, 1990). These preprocessing steps, which we call *classical pruning*, can significantly reduce the connectivity of the network, making accessible many queries that would be inaccessible otherwise. Although classical pruning can be very effective, one can identify situations where it does not exploit the full power of evidence, especially when the network contains local structure. The investigation in this paper serves to spotlight the power of evidence by discussing the extent to which it can be exploited computationally, and by introducing the notion of compiling a Bayesian network with evidence.

Traditionally, one incurs a compilation cost to prepare for answering a large number of queries over different evidence, amortizing the cost over the queries. But when evidence is fixed, this benefit may seem illusory at first. We will show, however, that compiling with evidence is often more tractable than compiling without evidence and that it can be very practical. First, the evidence may be fixed on only a subset of the variables, leaving room for posing queries with respect to other variables (this happens in MAP computations). Second, one may be interested in estimating the value of network parameters which will maximize the probability of given evidence (this happens, for example, in genetic linkage analysis). In this case, one may want to use iterative algorithms such as EM or gradient ascent (Dempster *et al.* , 1977), which pose many network queries with respect to the given evidence but different network parameter values. A similar application appears in sensitivity analysis, where the goal is to search for some network parameters that satisfy a given constraint.

Our approach to compiling with evidence is based on an approach to compiling a network without evidence into an arithmetic circuit (Darwiche, 2002; Darwiche, 2003); see Figure 1. The inputs to the circuit correspond to evidence indicators (for recording evidence) and network parameters and the output to the probability of recorded evidence under the given values of parameters. Given evidence, we will then compile an arithmetic circuit that is hardwired for that evidence and, hence, will only be good for computing queries with respect to that evidence. The particular compilation approach we adopt reduces the problem to one of logical inference, which we argue is a natural setting for exploiting the interaction between evidence and network local structure.

We apply the compilation approach to genetic linkage analysis, where we provide experimental results showing order of magnitude improvement over state of the art systems for certain benchmarks. We also show that the proposed approach subsumes empirically the quickscore algorithm (Heckerman, 1989).

This paper is organized as follows. Section 2 defines a semantics for compiling with evidence and describes areas where it applies. Section 3 describes our specific approach to compiling with evidence. We illustrate some of the reasons the approach works and apply it to genetic linkage analysis in Section 4. In Section 5, we show that the approach empirically subsumes the quickscore algorithm and applies to networks without determinism. Finally, Section 6 presents some concluding remarks.

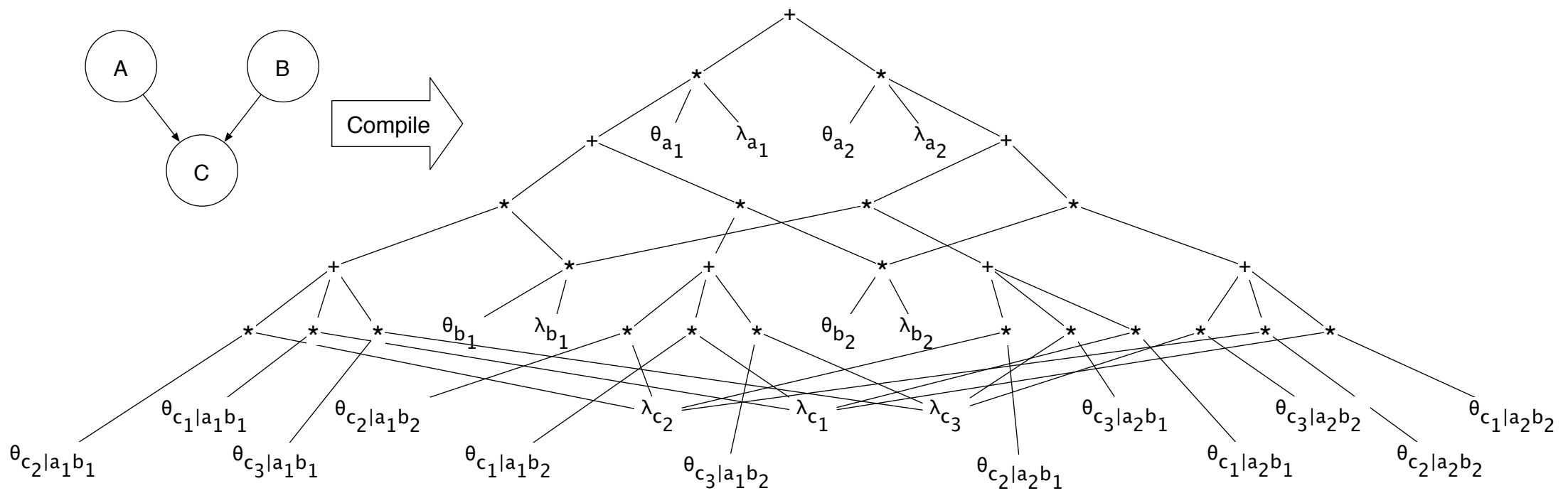

Figure 1: A Bayesian network and a corresponding AC.

# 2 COMPILING WITH EVIDENCE

This section defines a semantics for compiling a Bayesian network with evidence and explains some areas where such compilation can provide significant advantage.

## 2.1 Semantics

We begin by reviewing compilation without evidence as described in (Darwiche, 2003) and (Darwiche, 2002). We view each network as a multi–linear function (MLF), which contains two types of variables: an evidence indicator $\lambda_x$ for each value $x$ of each variable $X$ and a parameter variable $\theta_{x|\mathbf{u}}$ for each network parameter. The MLF contains a term for each instantiation of the network variables, and the term is the product of all indicators and parameters that are consistent with the instantiation. For example, consider the network in Figure 1, where $A$ and $B$ have two values, and $C$ has three values. The corresponding MLF involves twenty–three variables and contains twelve terms as follows,

$$\begin{aligned}
&\lambda_{a_1}\lambda_{b_1}\lambda_{c_1}\theta_{a_1}\theta_{b_1}\theta_{c_1|a_1,b_1} + \lambda_{a_1}\lambda_{b_1}\lambda_{c_2}\theta_{a_1}\theta_{b_1}\theta_{c_2|a_1,b_1} + \\
&\lambda_{a_1}\lambda_{b_1}\lambda_{c_3}\theta_{a_1}\theta_{b_1}\theta_{c_3|a_1,b_1} + \lambda_{a_1}\lambda_{b_2}\lambda_{c_1}\theta_{a_1}\theta_{b_2}\theta_{c_1|a_1,b_2} + \\
&\lambda_{a_1}\lambda_{b_2}\lambda_{c_2}\theta_{a_1}\theta_{b_2}\theta_{c_2|a_1,b_2} + \lambda_{a_1}\lambda_{b_2}\lambda_{c_3}\theta_{a_1}\theta_{b_2}\theta_{c_3|a_1,b_2} + \\
&\lambda_{a_2}\lambda_{b_1}\lambda_{c_1}\theta_{a_2}\theta_{b_1}\theta_{c_1|a_2,b_1} + \lambda_{a_2}\lambda_{b_1}\lambda_{c_2}\theta_{a_2}\theta_{b_1}\theta_{c_2|a_2,b_1} + \\
&\lambda_{a_2}\lambda_{b_1}\lambda_{c_3}\theta_{a_2}\theta_{b_1}\theta_{c_3|a_2,b_1} + \lambda_{a_2}\lambda_{b_2}\lambda_{c_1}\theta_{a_2}\theta_{b_2}\theta_{c_1|a_2,b_2} + \\
&\lambda_{a_2}\lambda_{b_2}\lambda_{c_2}\theta_{a_2}\theta_{b_2}\theta_{c_2|a_2,b_2} + \lambda_{a_2}\lambda_{b_2}\lambda_{c_3}\theta_{a_2}\theta_{b_2}\theta_{c_3|a_2,b_2}
\end{aligned} \quad (1)$$

To compute the probability of evidence $\mathbf{e}$, we evaluate the MLF after setting indicators that contradict $\mathbf{e}$ to 0 and other indicators to 1. For example, to compute $\mathsf{Pr}(a_2, b_1)$, we evaluate the above MLF after setting $\lambda_{a_1}$, $\lambda_{b_2}$ to 0 and $\lambda_{a_2}$, $\lambda_{b_1}$, $\lambda_{c_1}$, $\lambda_{c_2}$, $\lambda_{c_3}$ to 1. Setting indicators has the effect of excluding those terms that are incompatible with the evidence. Computing answers to other probabilistic queries, such as posterior marginals on network variables or families, can be obtained from the partial derivatives of the MLF; see (Darwiche, 2003) for details.

As is clear from the above description, the MLF has an exponential size. Yet, one may be able to factor this function and represent it more compactly. An *arithmetic circuit* (AC) is a rooted DAG, where each leaf represents a real–valued variable or constant and each internal node represents the product or sum of its children. Figure 1 depicts an AC. If we can factor the network MLF efficiently using an arithmetic circuit, then inference can be done in time linear in the size of the circuit, since the value and (first) partial derivatives of an arithmetic circuit can all be computed simultaneously in time linear in the circuit size (Darwiche, 2003). In an AC representing a network MLF, each leaf represents an indicator or parameter. An effective method of producing an AC is given in (Darwiche, 2002), and (Park & Darwiche, 2003b) shows that the AC is a generalization of the jointree.[1]

The MLF above and its corresponding AC are capable of answering queries with respect to *any* evidence. However, if we are willing to commit to specific evidence $\mathbf{e}$, then we can instead work with a much simpler MLF. For evidence $\mathbf{e} = \{a_2, b_1\}$, the above MLF can be reduced as follows,

$$\begin{aligned}
&\lambda_{c_1}\theta_{a_2}\theta_{b_1}\theta_{c_1|a_2,b_1} + \lambda_{c_2}\theta_{a_2}\theta_{b_1}\theta_{c_2|a_2,b_1} + \\
&\lambda_{c_3}\theta_{a_2}\theta_{b_1}\theta_{c_3|a_2,b_1}
\end{aligned} \quad (2)$$

[1]The AC is a generalization of Jointree in the sense that a Jointree embeds an AC with extra syntactic restrictions. Moreover, the two passes in jointree inference correspond to circuit evaluation and differentiation.

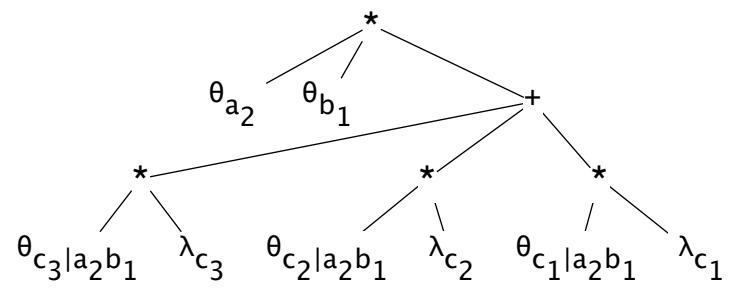

Figure 2: An AC that incorporates evidence.

In general, one obtains the *instantiated* MLF for a given network and evidence by removing each term from the MLF that contradicts the evidence. An instantiated MLF, and hence its corresponding AC, is therefore capable of answering only queries where $\mathbf{e}$ is given, such as $\mathsf{Pr}(\mathbf{e})$, $\mathsf{Pr}(X|\mathbf{e})$ for network variable $X$, and $\mathsf{Pr}(\mathbf{e}, \mathbf{m})$ for additional evidence $\mathbf{m}$. Figure 2 depicts the instantiated AC, which contrasts with the AC in Figure 1.

Although we have lost the ability to apply arbitrary evidence, compiling with evidence can be much more efficient than compiling without. Moreover, the instantiated AC still captures information that is critical to many inference tasks and does so in a way that provides significant advantage to an approach not based on compilation. We provide some examples next.

### 2.2 Applications

**Genetic linkage analysis** can model genetic information for a population of related individuals (a pedigree) using a Bayesian network (Fishelson & Geiger, 2002). Some network parameters $\theta_1, \ldots, \theta_n$ represent *recombination factors*, and the goal is to search for the recombination factors which maximize the probability of given evidence $\mathbf{e}$: $argmax_{\theta_1,\ldots,\theta_n} \mathsf{Pr}(\mathbf{e}|\theta_1, \ldots, \theta_n)$. The procedure amounts to ordering genes on a chromosome and determining the distance between them. Typically, one solves this problem by posing $\mathsf{Pr}(\mathbf{e}|\theta_1, \ldots, \theta_n)$ using particular values of recombination factors (parameters), and then repeating multiple times for different values. Our results demonstrate that, on some benchmarks, compilation can significantly improve on SUPERLINK 1.4,[2] which is a state–of–the–art system for the task.

**Sensitivity analysis** involves searching for parameter changes that satisfy certain constraints. For example, an expert may decide that $\mathsf{Pr}(A = a_1|\mathbf{e})$ must be greater than $\mathsf{Pr}(A = a_2|\mathbf{e})$ for some specific evidence $\mathbf{e}$. Our goal is to identify minimal parameter changes that satisfy this constraint. The problem can be solved relatively efficiently for a single parameter change, or multiple parameter changes within the same CPT (Chan & Darwiche, 2004). For multiple parameters spread over multiple CPTs, the solution involves numerical methods that pose multiple probabilistic queries under evidence $\mathbf{e}$, but with different values for network parameters. In this case, compiling the network with given evidence is quite practical, as the work done during compilation can be amortized over the many different queries.

**MAP** is the problem of computing the most likely instantiation of a set of variables $\mathbf{M}$ given evidence $\mathbf{e}$. Computing MAP can utilize compilation with evidence in a way that is similar to that of genetic linkage and sensitivity analysis, but instead of adjusting parameters, we adjust indicators. Both exact and approximate algorithms for computing MAP involve obtaining initial evidence $\mathbf{e}$ and then repeatedly computing $\mathsf{Pr}(\mathbf{e}, \mathbf{m})$ for different instantiations $\mathbf{m}$ of a subset of the MAP variables (Park & Darwiche, 2003a). We can therefore compile an AC with evidence $\mathbf{e}$ and then evaluate it for different values associated with indicators of variables $\mathbf{M}$.

## 3 IMPLEMENTATION

We now describe the technique used in the experimental results to compile a network with evidence into an AC. The approach for compiling a network without evidence into an AC has been described in (Darwiche, 2002), and is based on encoding the network MLF into a set of logical clauses (CNF), factoring the CNF, and then extracting an AC from the factored CNF. The details of factoring the CNF and extracting the AC are not critical here, so we will refer the reader to (Darwiche, 2002) for details. We will however review how a real–valued MLF can be encoded semantically into a propositional theory, and show how the network MLF can be encoded using a CNF. This is needed for explaining how evidence is exploited during compilation.

To illustrate the encoding scheme, consider the MLF $f = a + ad + abd + abcd$ over real–valued variables $a, b, c, d$. The basic idea is to specify this MLF using a propositional theory that has exactly four models, one for each term in $f$. Specifically, the propositional theory $\Delta_f = V_a \wedge (V_b \Rightarrow V_d) \wedge (V_c \Rightarrow V_b)$ over Boolean variables $V_a, V_b, V_c, V_d$ has exactly four models and encodes $f$ as follows,

| *Model* | $V_a$ | $V_b$ | $V_c$ | $V_d$ | *encoded term* |
|---|---|---|---|---|---|
| $\sigma_1$ | *true* | *false* | *false* | *false* | $a$ |
| $\sigma_2$ | *true* | *false* | *false* | *true* | $ad$ |
| $\sigma_3$ | *true* | *true* | *false* | *true* | $abd$ |
| $\sigma_4$ | *true* | *true* | *true* | *true* | $abcd$ |

That is, model $\sigma$ encodes term $t$ since $\sigma(V_j) = true$ iff term $t$ contains the real–valued variable $j$.

The encoding described above is semantic; that is, it describes the theory $\Delta_f$ which encodes a multi–linear function by describing its models. We specify these theories using a CNF that has one Boolean variable $V_\lambda$ for each indicator variable $\lambda$, and one Boolean variable $V_\theta$ for each parameter variable $\theta$. For brevity though, we will abuse notation and simply write $\lambda$ and $\theta$ instead of $V_\lambda$ and $V_\theta$. CNF clauses fall into three sets. First, for each network variable $X$ with domain $x_1, x_2, \ldots, x_n$, we have,

$$\underline{\text{Indicator clauses}}: \quad \lambda_{x_1} \vee \lambda_{x_2} \vee \ldots \vee \lambda_{x_n}$$
$$\neg\lambda_{x_i} \vee \neg\lambda_{x_j}, \text{ for } i < j$$

[2] http://bioinfo.cs.technion.ac.il/superlink/

These clauses ensure that exactly one of $X$'s indicator variables appears in each term of the MLF. The second two sets of clauses correspond to network parameters. In particular, for each parameter $\theta_{x_n|x_1,x_2,\ldots,x_{n-1}}$, we have,

$$\underline{\text{IP clause}}: \lambda_{x_1} \wedge \lambda_{x_2} \wedge \ldots \wedge \lambda_{x_n} \Rightarrow \theta_{x_n|x_1,x_2,\ldots,x_{n-1}}$$
$$\underline{\text{PI clauses}}: \theta_{x_n|x_1,x_2,\ldots,x_{n-1}} \Rightarrow \lambda_{x_i}, \text{ for each } i$$

The models of this CNF are in one–to–one correspondence with the terms of the MLF. In particular, each model of the CNF will correspond to a unique network variable instantiation, and will set to true only those indicator and parameter variables which are compatible with that instantiation. The encoding we use in our experiments is a bit more sophisticated than described above (Chavira & Darwiche, 2005), but the above encoding will suffice to make our points below.

The encoding as discussed does not include information about evidence. Recall that to incorporate evidence $\mathbf{e}$, we need to exclude MLF terms that contradict $\mathbf{e}$. It is quite easy to do so in the current framework. Consider the network in Figure 1, its MLF (1), and the evidence $\{a_2, b_1\}$. Assume that we have generated the CNF $\Delta$ for this network. Our goal becomes excluding from $\Delta$ models corresponding to terms that contradict the evidence. We can easily do so by adding the following unit clauses to the CNF: $\lambda_{a_2}$ and $\lambda_{b_1}$. In general, to incorporate evidence $x_1$, $x_2$, ..., $x_n$, we add unit clauses $\lambda_{x_1}$, $\lambda_{x_2}$, ..., $\lambda_{x_n}$. Moreover, it is easy to incorporate more general types of evidence. For example, we can incorporate the assertion "$c_1$ or ($a_1$ and $b_2$)" by including the clauses $\lambda_{c_1} \vee \lambda_{a_1}$ and $\lambda_{c_1} \vee \lambda_{b_2}$.

In our implementation, we simplify the constructed CNF together with evidence by running unit resolution and then removing subsumed clauses. We then invoke our compiler which factors the CNF using a version of the recursive conditioning algorithm (Darwiche, 2004). This algorithm makes repeated use of conditioning to decompose the CNF into disconnected CNFs that are compiled independently. Moreover, the algorithm runs unit resolution after each conditioning to simplify the CNF further. This process of decomposition becomes much more effective given the initial evidence injected into the CNF, which helps to simplify the CNF considerably. Some of the benefit is obtained immediately from the initial preprocessing of the CNF. Other benefits, however, are obtained during the compilation process itself since conditioning sets the value of variables, which together with the injected evidence can lead to even more simplification of the CNFs and, hence, better decomposition. We will see examples of this behavior in the following section.

# 4 THE POWER OF EVIDENCE

Consider the "Original Evidence" portion of Table 1, which contains a set of Bayesian networks corresponding to pedigrees in the domain of genetic linkage analysis. Each network has been classically pruned for specific evidence, yet they still have very connected topologies, as shown by the cluster sizes obtained using a minfill variable ordering heuristic.[3] None of these networks could be compiled without evidence, yet the table lists data on successful compilations for most of these networks once evidence is introduced, despite the large cluster sizes.[4] In particular, the table shows the offline time (which includes preprocessing and compiling), size of AC, and online inference time for computing $\mathsf{Pr}(\mathbf{e})$. Note that online inference may be repeated for new recombination values, without re–incurring the offline cost.

Our current implementation uses only unit resolution and removal of subsumed clauses during its simplification of the CNF before compiling. However, based on the amount of determinism in these networks, more advanced logical techniques can be utilized. We therefore augmented the given evidence with some additional evidence learned by the domain specific Lange and Goradia algorithm (Lange & Goradia, 1987). It should be noted that this additional evidence can be inferred by standard logical techniques applied to the initial evidence and network determinism, and could therefore be made domain independent. By using this additional (inferred) evidence, we can see in the "Learned Evidence" portion of Table 1 that all these networks compile in a reasonable amount of time, and that online inference is faster. Since the additional learned evidence may apply to internal (non-leaf) nodes, one may use this evidence to empower classical pruning. Indeed the table lists the adjusted cluster sizes for these networks after having applied classical pruning using the additional evidence. The learned evidence makes many of these networks accessible to classical

[3]We are reporting here normalized cluster sizes (log 2 of the number of instantiations of a given cluster).

[4]Experiments in this paper ran on a 1.6GHz Pentium M processor with 2GB of memory.

Table 1: EA results.

| | Original Evidence | | | | Learned Evidence | | | | Full Preprocessing | | | |
|---|---|---|---|---|---|---|---|---|---|---|---|---|
| NET | MAX CLUST | OFFLINE SEC | AC EDGES | Pr(e) SEC | MAX CLUST | OFFLINE SEC | AC EDGES | Pr(e) SEC | MAX CLUST | OFFLINE SEC | AC EDGES | Pr(e) SEC |
| ea1 | 31.6 | 2.67 | 98,613 | 0.01 | 13.0 | 1.57 | 24,055 | 0.00 | 11.3 | 1.30 | 20,230 | 0.01 |
| ea2 | 38.6 | 3.65 | 144,181 | 0.01 | 13.0 | 1.57 | 28,390 | 0.01 | 11.3 | 1.35 | 21,218 | 0.01 |
| ea3 | 40.6 | 10.45 | 272,503 | 0.01 | 13.0 | 1.68 | 31,575 | 0.01 | 11.3 | 1.34 | 20,489 | 0.01 |
| ea4 | 46.0 | 8.02 | 322,063 | 0.03 | 13.0 | 1.89 | 34,126 | 0.01 | 11.3 | 1.42 | 19,455 | 0.01 |
| ea5 | 60.9 | 38.48 | 992,917 | 0.03 | 13.0 | 2.79 | 54,703 | 0.01 | 11.3 | 1.64 | 22,963 | 0.01 |
| ea6 | 70.9 | 125.62 | 3,557,015 | 0.11 | 14.9 | 7.75 | 120,700 | 0.01 | 12.3 | 1.96 | 31,146 | 0.01 |
| ea7 | 82.9 | 4,591.65 | 26,934,471 | 3.35 | 15.6 | 73.20 | 997,652 | 0.03 | 12.3 | 2.34 | 39,957 | 0.02 |
| ea8 | 106.9 | 1,732.10 | 24,375,244 | 6.05 | 16.0 | 9.35 | 180,100 | 0.01 | 12.3 | 3.24 | 41,249 | 0.01 |
| ea9 | 200.5 | n/a | n/a | n/a | 28.3 | 1,226.43 | 3,597,965 | 1.81 | 12.3 | 7.50 | 82,297 | 0.03 |
| ea10 | 204.1 | n/a | n/a | n/a | 31.6 | 1,665.57 | 13,758,985 | 1.42 | 12.3 | 8.61 | 95,417 | 0.03 |
| ea11 | 235.1 | n/a | n/a | n/a | 30.6 | 2,586.64 | 12,298,513 | 5.81 | 12.3 | 10.04 | 92,274 | 0.03 |

Table 2: EE results with full preprocessing.

| NET | MAX CLUST | OFFLINE SEC | AC EDGES | Pr(e) SEC | SUPERLINK SEC |
|---|---|---|---|---|---|
| ee33 | 20.2 | 25.33 | 2,070,707 | 0.59 | 1046.72 |
| ee37 | 29.6 | 61.29 | 1,855,410 | 0.39 | 1381.61 |
| ee30 | 35.9 | 376.78 | 27,997,686 | 8.37 | 815.33 |
| ee23 | 38.0 | 89.47 | 3,986,816 | 1.08 | 502.02 |
| ee18 | 41.5 | 283.96 | 23,632,200 | 6.63 | 248.11 |

inference algorithms, but three of the networks still pose problems for classical techniques.

It is worth putting the results in perspective by comparing to state-of-the-art results in genetic linkage analysis obtained with SUPERLINK. This system uses a combination of variable elimination and conditioning, along with many domain specific preprocessing rules and a sophisticated search for a variable ordering. All SUPERLINK timings we report include preprocessing and computing answers to two $\mathsf{Pr}(\mathbf{e})$ queries, where on difficult networks, the majority of the time is spent doing inference. Until the latest release, the networks in Table 1 were considered very challenging, with EA11 taking over 10 hours. The newest version of SUPERLINK, 1.4, includes enhancements that preprocess and perform the two $\mathsf{Pr}(\mathbf{e})$ queries in 7 seconds on even the most difficult of these networks. If we allow ourselves to use further simplification techniques, which include some simplifications from SUPERLINK 1.4 and also some rules to detect variable equivalence, we obtain the results shown in the "Full Preprocessing" portion of Table 1. Here, offline time takes about 10 seconds on the hardest network and online inference is very fast.

More dramatic are the results reported in Table 2 on five networks from the challenging SUPERLINK 1.4 data sets. On these networks, the compilation approach was able to improve on SUPERLINK's performance as reported in (Fishelson *et al.* , 2004). On four of these networks, offline time is shorter than the SUPERLINK time. Once compiled, the generated ACs can repeatedly compute $\mathsf{Pr}(\mathbf{e})$ extremely efficiently compared to SUPERLINK. Because one of the main tasks of genetic linkage analysis is to do maximum likelihood estimation over many iterations, the ability to perform online inference quickly is critical.

Note that we can differentiate these circuits and, hence, obtain marginals over families in about $2-3$ times as long as it takes to evaluate $\mathsf{Pr}(\mathbf{e})$. This allows us to run the EM algorithm, which requires marginals over families to perform each iteration. When comparing these timings with the time it would take to re-run SUPERLINK for the same purpose, one sees the significant benefit of compiling with evidence. Suppose for example that we have 10 parameters we want to estimate and that EM or gradient ascent takes 20 iterations to converge. For network ee33, we would perform 200 queries in about 350 seconds using AC, whereas running SUPERLINK to compute those values would require days.

### 4.1 Examples

We now demonstrate how combining evidence with local structure can make the inference process more efficient. These gains cannot be obtained using classical pruning, although some can be obtained using more sophisticated schemes (e.g., (Larkin & Dechter, 2003)).

The first example uses the network in Figure 1, where $A$ and $B$ have two states and $C$ has three states. Let the CPT for the variable $C$ contain all zeros except for the four lines below.

| $A$ | $B$ | $C$ | $Pr(C\|A,B)$ |
|---|---|---|---|
| $a_1$ | $b_1$ | $c_1$ | 1.0 |
| $a_1$ | $b_2$ | $c_2$ | 1.0 |
| $a_2$ | $b_1$ | $c_2$ | 1.0 |
| $a_2$ | $b_2$ | $c_3$ | 1.0 |

Suppose we know that $C = c_1$. From this information, we can logically infer $A = a_1$ and $B = b_1$. In fact, this information can be obtained by preprocessing the CNF encoding of the network using unit resolution. The learned evidence can be added to the CNF, or it could be used to empower classical pruning.

Now assume that we have evidence $\{c_2\}$. Because $A$ and $B$ are binary, there would normally be four possible configurations of $A$ and $B$. However, given the CPT parameterization, we can rule out both $\{a_1, b_1\}$ and $\{a_2, b_2\}$, leaving only two configurations. This conclusion can again be obtained by applying unit resolution to our CNF encoding. However, in this case, the inferred information cannot be expressed in terms of classical pruning. Furthermore, it cannot be expressed using a more advanced form of simplification, where variable states known to never occur are removed form the variable's domain, since every state of $A$ and $B$ is valid. The learned "multi-variable" evidence is, however, easily written in the language of CNF, and can be utilized in further simplifications during the compilation process.

One question is how often situations like the above occur in real networks. The examples actually derive from real networks in the domain of genetic linkage analysis, where variables $A$ and $B$ represent a person's genotype (for example a person's blood type) and $C$ represents the phenotype (the observed portion of the genotype). The example then shows one way the genotype can be mapped to the phenotype. Take the simplified case where there are only two blood types, 1 and 2. Then the four possible genotype combinations are 1/1, 1/2, 2/1, 2/2, although frequently 1/2 and 2/1 cannot be differentiated, so there are only three phenotypes. The example models this situation by mapping two configurations of $A$ and $B$ to the same value for $C$. Furthermore, in this domain, evidence is typically on phenotype variables, which translates to evidence on $C$ in our model.

The third example from genetic linkage analysis involves four variables: child $C$ with parents $A, B$, and $S$. The variable $C$ in this case is not the phenotype, but the genotype in a child which is inherited from one of the parent's genes, $A/B$, based on the value of $S$. We assume that all four variables are binary and that the portion of the table with $S = s_1$ is as follows,[5]

| $S$ | $A$ | $B$ | $C$ | $Pr(C\|A, B)$ |
|---|---|---|---|---|
| $s_1$ | $a_1$ | $b_1$ | $c_1$ | 1.0 |
| $s_1$ | $a_1$ | $b_2$ | $c_1$ | 1.0 |
| $s_1$ | $a_2$ | $b_1$ | $c_2$ | 1.0 |
| $s_1$ | $a_2$ | $b_2$ | $c_2$ | 1.0 |

The point of this example is to illustrate how compilation can utilize evidence even when preprocessing cannot. This type of gain is one of the factors that allows us to successfully compile a network whose treewidth remains high after preprocessing. Compiling repeatedly conditions to decompose the CNF. Let us consider the case where we are given evidence $\{c_1\}$, and during compilation, we condition on $S = s_1$. Assuming a proper encoding of the network into CNF, combining the evidence with the value for $S$ allows us to infer $a_1$, which unit resolution can use to achieve further gains. Conditioning on $S = s_2$ yields a similar conclusion for $b_1$. In this case, the full power of evidence on $C$ is realized only when combined with conditioning, which takes place during the compilation step.

[5]In general, the variables are multi–valued, and this discussion also applies in this case.

Table 3: Friends and Smokers Results.

| DOM SIZE | MAX CLUST | OFFLINE SEC | AC EDGES | ONLINE SEC |
|---|---|---|---|---|
| 1 | 3 | 0.02 | 18 | 0.00 |
| 4 | 13 | 0.03 | 293 | 0.01 |
| 7 | 36 | 0.08 | 1,295 | 0.01 |
| 10 | 70 | 0.34 | 3,512 | 0.02 |
| 13 | 118 | 1.07 | 7,430 | 0.03 |
| 16 | 172 | 3.21 | 13,535 | 0.04 |
| 19 | 244 | 9.04 | 22,313 | 0.05 |
| 22 | 316 | 23.56 | 34,250 | 0.07 |
| 25 | 412 | 48.32 | 49,832 | 0.09 |
| 28 | 528 | 105.74 | 69,545 | 0.13 |
| 29 | 560 | 130.00 | 77,118 | 0.14 |

We close this section by quickly examining one more set of networks. (Richardson & Domingos, 2004) discusses a relational probabilistic model involving an imaginary domain of people and relations on the people including which smoke, which have cancer, and who are friends of whom. There are also logical constraints on the model, such as the constraint that if a person's friend smokes, then the person also smokes. We worked with a slight variation on this model, and each network in Table 3 represents an instance for a different number of people. For a given network, some nodes represent ground relations and others represent logical constraints. The key point is that, in the absence of evidence, we could only compile the first two networks listed. However, when we commit to evidence asserting that the logical constraints are *true*, the networks become relatively easy, the hardest requiring 130 seconds to compile and 0.14 seconds for online inference. Online inference involves asserting evidence $\mathbf{e}$ on some of the relations and computing $\mathsf{Pr}(\mathbf{e})$ and marginals on all remaining relations.

## 5 THE QUICKSCORE ALGORITHM

We illustrate two points in this section. First, our compiling approach empirically subsumes the quickscore algorithm, a dedicated algorithm for two–level noisy–or networks. Second, networks which do not contain determinism, and hence may not look amenable to exploiting evidence as described earlier, can be transformed to make them amenable to these techniques.

We start by considering two–level noisy–or networks as given in Figure 3(a). Here, each $d_i$ represents a dis-ease, which occurs in a patient with probability $p_i$, and each $f_j$ represents a feature (e.g., test result), which we may observe to be negative or positive in the patient. We assume a noisy–or relationship between a specific feature $f_j$ and the diseases $d_i$ that may causes it. That is, if $d_i$ is not present, then $d_i$ will not cause $f_j$. Otherwise, $d_i$ will cause $f_j$ with probability $p_{i,j}$. We wish to compute a marginal for each disease given evidence on features. Standard inference has a worst case time complexity that is exponential in the number $n$ of diseases. However, (Heckerman, 1989) showed that computing such marginals can be done in time exponential only in the size $m^+$ of the set $F^+$ of features known to be positive. The argument makes several appeals to the independence relationships created by the noisy–or assumptions and by the network structure. It culminates with the definition of the quickscore algorithm, which iterates over the power set of $F^+$ and computes a marginal on a single disease in time $\Theta(nm^- 2^{m^+})$, where $m^-$ is the number of negative findings.

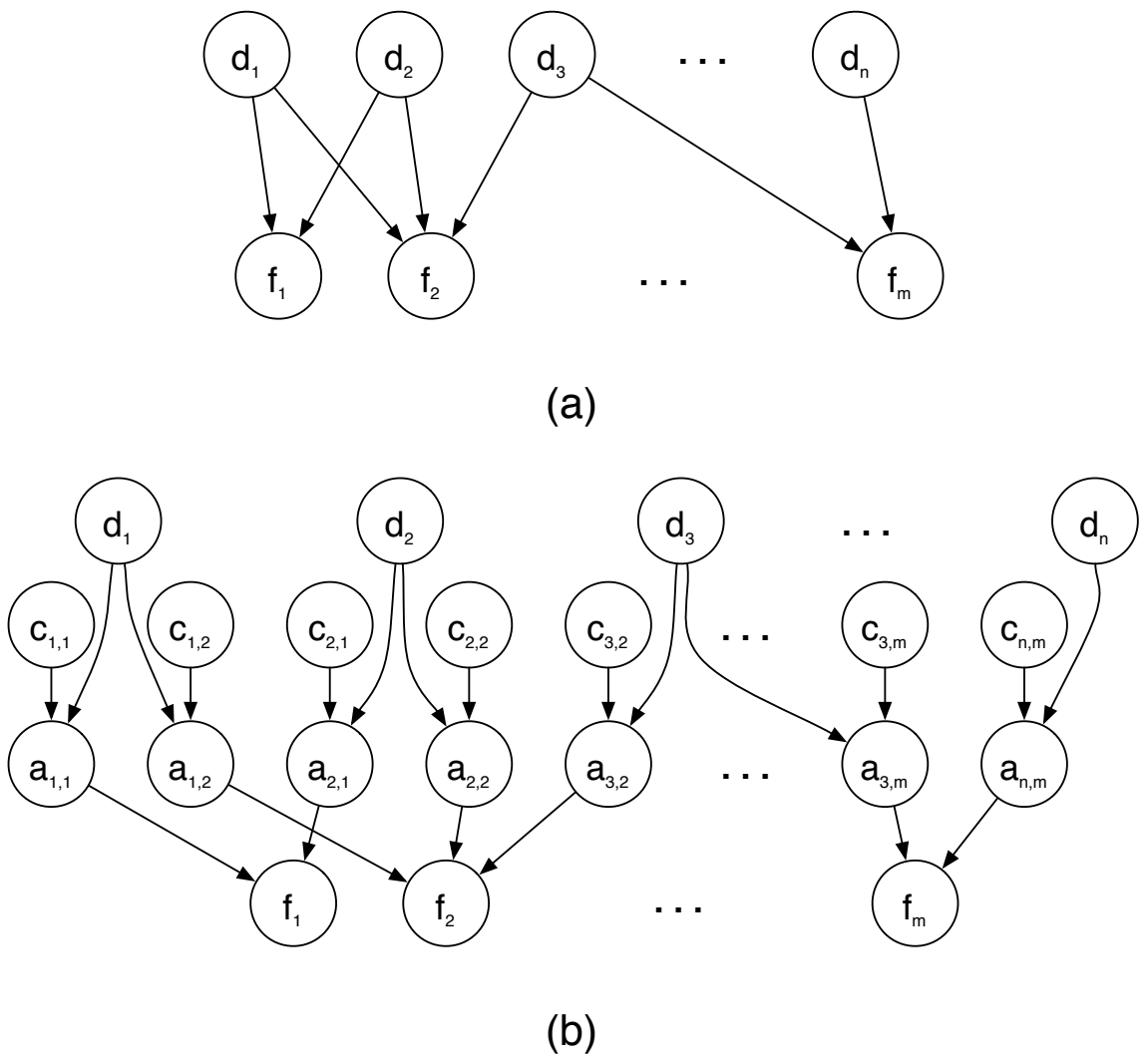

Figure 3: (a) A disease/feature network; (b) the network with determinism introduced.

It is well–known that the semantics of the noisy–or relationships allows us to transform the network in Figure 3(a) into the network in Figure 3(b). Here, each edge from $d_i$ to $f_j$ in the original network is replaced with two nodes, $c_{i,j}$ and $a_{i,j}$, and three edges. Each introduced $c_{i,j}$ is a binary root such that $\mathsf{Pr}(c_{i,j}) = p_{i,j}$; and each introduced $a_{i,j}$ represents whether disease $d_i$ causes feature $f_j$. Therefore, $a_{i,j}$ represents a conjunction of $d_i$ and $c_{i,j}$, and each feature is a disjunction of its parents. This disjunction can be represented very compactly in CNF, even when there are a large number of parents. Although the network in Figure 3(a) typically does not possess determinism, the transformed network possesses an abundance in the form of introduced conjunctions and disjunctions, leading one to wonder whether combining this determinism with evidence in the manner proposed would duplicate quickscore's performance. To test this hypothesis, we chose different values for $m^+$, and for each, we constructed ten experiments, each designed to be similar to the experiments on the proprietary network used to demonstrate quickscore. For each experiment, we generated a random network containing 600 diseases and 4100 features. For each feature, we chose 11 possible causes uniformly from the set of diseases. We then chose each $p_i$ and each $p_{i,j}$ uniformly from the open interval $(0, 1)$. In addition, we generated evidence by setting to positive $m^+$ features chosen uniformly from the set of features and setting the remaining features to negative.[6] In this way, the experiment utilizes its own randomly generated network and its own randomly generated evidence. Finally, we compiled and evaluated the network with the evidence, yielding a marginal over each disease.

Each of the experiments produced a network for which minfill computed a maximum cluster size between 586 and 589. Because the set of evidence variables is the same as the set of leaves in the network, classical pruning would have no effect on this cluster size.

For each value of $m^+$, Table 4 shows results, averaged over the ten experiments. The most important observation is that the approach to compiling with evidence empirically subsumes quickscore. Indeed, quickscore is exponential in $m^+$ even in the best case, whereas compiling was sometimes fast, even for large $m^+$. For example, the minimum compile times for $m^+ = 28$ and $m^+ = 29$ were $41s$ and $135s$, respectively. Furthermore, quickscore computes a marginal only for a single disease, whereas the described method computes marginals over all 600 diseases simultaneously.

The transformation to introduce determinism applies not only to the types of networks on which quickscore runs, but to any network involving noisy–or relationships. There are similar transformations for other types of local structure, including noisy–or with leak, noisy–max (Díez & Galán, 2003), and context–specific independence (Boutilier *et al.* , 1996). Consider a final example involving a family containing binary variable $C$ with binary parents $A$ and $B$. Suppose that given $A = a_1$, $C$ becomes independent of $B$; yet this is not true for $A = a_2$. In this case, we introduce auxiliary variable $S$ with three states between $A/B$ and $C$. $S$'s CPT is deterministic and sets $S$ to $s_1$ when $A = a_1$, to $s_2$ for parent state $\{a_2, b_1\}$, and to $s_3$ for parent state $\{a_2, b_2\}$. Moreover, the CPT for $C$ becomes,

[6] We could have left some features $F'$ out of the evidence. In this case, classical pruning would suffice to remove nodes $F'$ from the network.

Table 4: Averaged diagnosis results.

| TRUE FEATURES | OFFLINE SEC | AC EDGES | ONLINE SEC |
|---|---|---|---|
| 0 | 23.73 | 48,100 | 0.05 |
| 3 | 23.86 | 52,830 | 0.05 |
| 6 | 23.81 | 57,638 | 0.05 |
| 9 | 23.82 | 62,547 | 0.05 |
| 12 | 24.19 | 67,632 | 0.05 |
| 15 | 23.60 | 73,321 | 0.04 |
| 18 | 24.95 | 81,629 | 0.05 |
| 21 | 30.95 | 109,335 | 0.05 |
| 23 | 42.81 | 145,333 | 0.06 |
| 25 | 155.12 | 434,445 | 0.08 |
| 27 | 469.70 | 1,141,674 | 0.17 |
| 28 | 728.52 | 1,691,833 | 0.23 |
| 29 | 1,046.93 | 2,352,820 | 0.30 |

| $S$ | $Pr(c_1 \mid S)$ |
|---|---|
| $s_1$ | $Pr(c_1 \mid a_1, b_1) = Pr(c_1 \mid a_1, b_2)$ |
| $s_2$ | $Pr(c_1 \mid a_2, b_1)$ |
| $s_3$ | $Pr(c_1 \mid a_2, b_2)$ |

Given evidence $A = a_1$, our logic–based strategy can infer both the value of $S$ and the independence of $C$ from $B$. This technique allows for more efficient decomposition during the compilation process, even though the original network contains no determinism.

# 6 CONCLUSION

We discussed the exploitation of evidence in probabilistic inference and highlighted the extent to which it can render inference tractable. We proposed a particular notion and approach for compiling networks with evidence, and discussed a number of practical applications to maximum likelihood estimation, sensitivity analysis and MAP computations. We presented several empirical results illustrating the power of proposed approach, and showed in particular how it empirically appears to subsume the performance of the quickscore algorithm.

## References

Boutilier, Craig, Friedman, Nir, Goldszmidt, Moises, & Koller, Daphne. 1996. Context-specific independence in Bayesian networks. *Proceedings of UAI-1996, 115–123.*

Chan, Hei, & Darwiche, Adnan. 2004. Sensitivity Analysis in Bayesian Networks: From Single to Multiple Parameters. *Proceedings of UAI-2004, 67–75.*

Chavira, Mark, & Darwiche, Adnan. 2005. Compiling bayesian networks with local structure. *IJCAI-2005.*

Darwiche, Adnan. 2002. A logical approach to factoring belief networks. *Proceedings of KR, 409–420.*

Darwiche, Adnan. 2003. A differential approach to inference in bayesian networks. *Journal of the ACM*, **50**(3), 280–305.

Darwiche, Adnan. 2004. New advances in compiling CNF to decomposable negational normal form. *Proceedings of European conference on artificial intelligence.*

Dempster, A., Laird, N., & Rubin, D. 1977. Maximum likelihood from incomplete data via the EM algorithm. *Journal of the royal statistical society*, 1–38.

Díez, Francisco & Galán, Severino. 2003. Efficient computation for the noisy max. *International Journal of Intelligent Systems*, **18**, 165–177.

Fishelson, Ma'ayan, & Geiger, Dan. 2002. Exact genetic linkage computations for general pedigrees. *Bioinformatics*, **18**(1), 189–198.

Fishelson, Ma'ayan, Dovgolevsky, Nickolay, & Geiger, Dan. 2004. *Maximum likelihood haplotyping for general pedigrees.* Tech. rept. CS-2004-13. Technion, Haifa, Israel.

Heckerman, David. 1989. A tractable inference algorithm for diagnosing multiple diseases. *Proceedings of UAI-1989, 174–181.*

Lange, K, & Goradia, TM. 1987. An algorithm for automatic genotype elimination. *American journal of human genetics*, **40**, 250–256.

Larkin, David, & Dechter, Rina. 2003. Bayesian inference in the presence of determinism. *AI and Statistics (AI-STAT).*

Park, J., & Darwiche, A. 2003a. Solving map exactly using systematic search. *Proceedings of UAI-2003, 459–468.*

Park, James, & Darwiche, Adnan. 2003b. A differential semantics for jointree algorithms. *Advances in neural information processing systems 15*, vol. 1, 299–307.

Richardson, Matt, & Domingos, Pedro. 2004. *Markov logic networks.* Tech. rept. Department of Computer Science, University of Washington.

Shachter, Ross D. 1986. Evaluating influence diagrams. *Operations research*, **34**(6), 871–882.

Shachter, Ross D. 1986. Evidence Absorption and Propagation Through Evidence Reversals. *Proceedings of UAI-1990.*